\documentclass[10pt,conference]{IEEEtran}

\usepackage{graphicx}
\usepackage{listings}
\usepackage{wrapfig}
\usepackage{caption}
\usepackage{booktabs}
\usepackage{tabularx}
\usepackage{float}
\usepackage{subcaption}
\usepackage{lscape}
\usepackage[numbers]{natbib}
\usepackage{multirow}
\usepackage{multicol}
\usepackage{url}
\usepackage{amsmath}
\usepackage{ragged2e}
\usepackage{color}
\usepackage{tabularray}
\usepackage{xcolor}
\lstdefinestyle{langstyle}{
  basicstyle=\ttfamily\footnotesize,
  keywordstyle=\color{blue},
  commentstyle=\color{dkred},
  stringstyle=\color{dkgreen},
  keepspaces=true,              
  breaklines=true,
  otherkeywords={::=},
  numberstyle=\ttfamily\footnotesize\color{gray},
  stepnumber=1,
  numbersep=8pt,
  backgroundcolor=\color{white},
  tabsize=4,
  showspaces=false,
  showstringspaces=false,
  xleftmargin=.18in,
  captionpos=b,
  escapeinside={(?}{?)},
  frame = single,
}
\definecolor{dkgreen}{rgb}{0,0.5,0}
\definecolor{dkred}{rgb}{0.5,0,0}
\definecolor{gray}{rgb}{0.5,0.5,0.5}
\definecolor{vlgray}{gray}{0.95}
\definecolor{lgray}{gray}{0.7}
\definecolor{bluehighlight}{HTML}{46adb7}
\definecolor{orangehighlight}{HTML}{e3a24d}
\definecolor{redhighlight}{HTML}{b95e8c}
\definecolor{codegray}{gray}{0.9}
\definecolor{Silver}{rgb}{0.752,0.752,0.752}
\definecolor{Gallery}{rgb}{0.937,0.937,0.937}
\definecolor{SwansDown}{rgb}{0.819,0.933,0.89}
\definecolor{HawkesBlue}{rgb}{0.854,0.925,0.984}
\definecolor{PastelPink}{rgb}{1,0.839,0.882}
\definecolor{FringyFlower}{rgb}{0.709,0.89,0.819}
\definecolor{TropicalBlue}{rgb}{0.784,0.89,0.976}
\definecolor{Lavenderblush}{rgb}{1,0.898,0.925}
\definecolor{PuertoRico}{rgb}{0.356,0.76,0.607}
\definecolor{JordyBlue}{rgb}{0.494,0.741,0.945}
\definecolor{PinkSalmon}{rgb}{1,0.521,0.647}
\definecolor{MonteCarlo}{rgb}{0.474,0.803,0.678}
\definecolor{Perano}{rgb}{0.639,0.815,0.96}
\definecolor{CarnationPink}{rgb}{1,0.678,0.764}

\usepackage[capitalize]{cleveref}
\newcommand{\added}[1]{{\color{black} #1}}

\AtBeginDocument{%
  \providecommand\BibTeX{{%
    \normalfont B\kern-0.5em{\scshape i\kern-0.25em b}\kern-0.8em\TeX}}}

\usepackage{subfiles} 

\begin{document}

\title{Skill over Scale: The Case for Medium, Domain-Specific Models for SE\vspace{-5pt}}











\author{
    \IEEEauthorblockN{Manisha Mukherjee, Vincent J. Hellendoorn}
    \IEEEauthorblockA{Carnegie Mellon University, Pittsburgh, USA\\
    \{mmukherj, vhellendoorn\}@cmu.edu}
}
\vspace{-20pt}



\maketitle
\IEEEpeerreviewmaketitle

\begin{abstract}

Recent advancements in AI have sparked a trend in constructing large, generalist language models that handle a multitude of tasks, including many code-related ones. While these models are expensive to train and are often closed-source, they have enjoyed broad adoption because they tend to outperform smaller, domain-specific models of code. In this work, we argue that this is not a foregone conclusion. We show that modestly sized domain-specific models can outperform much larger ones on code labeling tasks, provided they are trained to the same standards. Concretely, we focus on StackOverflow (SO), which offers large volumes of aligned code and text data. We align established best-practices for pre-training large language models with properties of StackOverflow as a data source, especially using a large context window (2,048 tokens), coupled with a powerful toolkit (Megatron-LM) to train two models: SOBertBase\cite{SOBertBase}, with 125M parameters, and SOBertLarge\cite{SOBertLarge} with 762M parameters, at a budget of just \added{\$374} and \added{\$1600} each. We compare the performance of our models with a prior domain-specific model which did not adopt many of these practices (BERTOverflow), as well two general-purpose BERT models (BERTBase and BERTLarge), and two models in OpenAI's GPT series (GPT-3.5 and GPT-4). We study four labeling tasks: question quality prediction, closed question prediction, named entity recognition and obsoletion prediction. The final task is a new benchmark we introduce, on which we additionally compare SOBert with a fine-tuned CodeLlama and StackLlama (models with 10x more parameters than SOBertLarge). Our models, including the smaller one, consistently outperform all baselines. In contrast, BertOverflow is outperformed by generalist models in most tasks. These results demonstrate that pre-training both \textit{extensively} and \textit{properly} on in-domain data can yield a powerful and affordable alternative to leveraging closed-source general-purpose models. Both models are released to the public with over 500 downloads in the last month alone on Hugging Face.
\end{abstract}


\section{Introduction}
\begin{figure}
\includegraphics[width=0.5\textwidth]{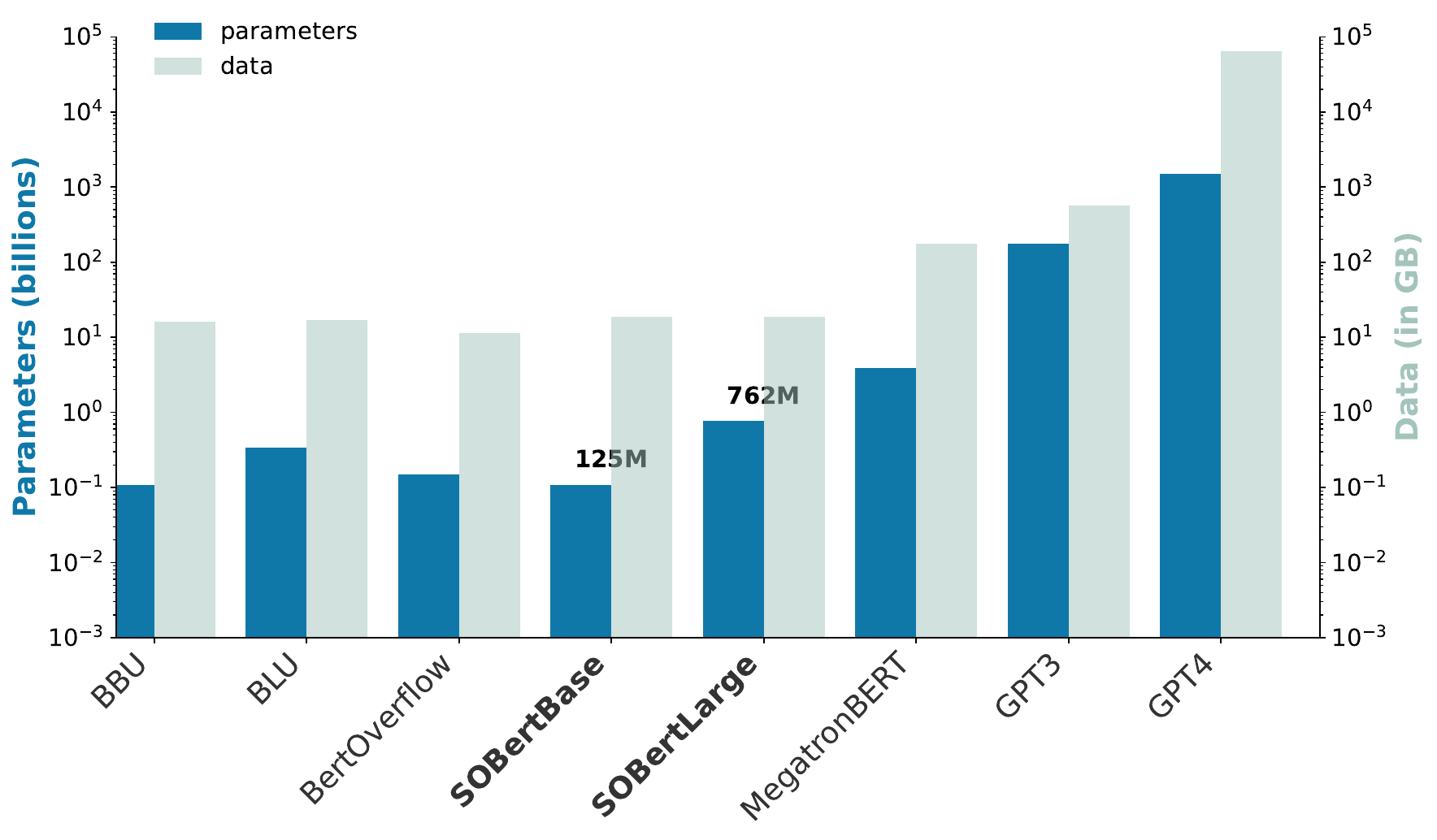}
\caption{Model and dataset sizes of state-of-the-art LLMs.}
\vspace{-15pt}
\label{fig:comparison}
\end{figure}

The software development process is changing rapidly, driven by advances in neural Language Models (LMs). While LMs can be trained from scratch on any given dataset, it has been found especially effective to \emph{pre-train} models on large amounts of data to learn general patterns from text or code \cite{dontStopPretraining}, and then fine-tune or instruction-tune models for downstream usage \cite{devlin2018bert, brown2020language}. The largest such models, referred to as Large Language Models (LLMs),
consistently improve in performance with larger datasets and more parameters \cite{hoffmann2022training}. \emph{Generalist models}, such as GPT-4 \cite{openai2024gpt4}, therefore train on datasets spanning many languages, domains, and modalities, including source code. These are expensive to train and often closed-source \cite{hellendoorn2021growing}, but broadly popular because of their strong performance across many domains.
Figure \ref{fig:comparison} shows the relative scale of various language models, including ours (in bold) compared to GPT-4 (note the log-scale on both axes).\footnote{Where dataset size is only reported in terms of tokens, we assume an average token size of 5 bytes. GPT-4 sizes are based 
on unpublished, but widely repeated claims.}

Using code data during pretraining enables LLMs to model code as just another form of language, with significant success \cite{chen2021evaluating}. Consequently, LLMs have found many applications in software engineering (SE), including in code completion \cite{liu2020multi}, bug localization \cite{ciborowska2022fast}, and code summarization \cite{wang2020fret}.
This has created a downward pressure on task- and domain-specific models that were previously popular, such as CodeT5 and CodeBERT \cite{wang2021codet5, feng2020codebert}.
In particular, many code-specific resources, including StackOverflow, contain ``just" billions of tokens -- far too little to train even modest LLMs. It would seem that generalist models are the way forward.

Yet the story is not over for domain-specific models of code. While LLMs thrive on generative tasks, they offer less impressive performance on classification tasks in specialized domains. Smaller, BERT-style models thrive on these in large part because they can be fine-tuned easily. LLMs are almost exclusively tuned to new tasks via prompting. In our experiments, for instance, BertOverflow \cite{tabassum2020code}, a BERT model trained specifically on StackOverflow content, outperforms GPT-4 on two out of three tasks where they are compared. Surprisingly, though, two generic (data-generalist but small) BERT models (BBU and BLU) mostly outperform GPT-4 \underline{and} BertOverflow. That might seem to imply that generalist models are still better, as long as fine-tuning is employed, but we conjecture otherwise.

To properly establish whether generalist models should be preferred over domain-specific ones in SE, we need to ensure that the latter are given all the same methodological advantages as the former. The ML community has spent considerable effort developing robust training approaches for LLMs, including extensive scaling laws \cite{kaplan2020scaling} that relate available compute to ideal model parameter, training configuration, and dataset sizes. This has resulted in training frameworks such as MegatronLM \cite{shoeybi2019megatron} that enable robust training of LMs with, among others, large batch sizes and context windows. We adopt these insights in our work.
Going back to the previous case, BBU uses only slightly more data than BertOverflow, but presents that data to the model in chunks of up to 512 tokens. BertOverflow, meanwhile, is trained with sentence-level inputs. StackOverflow posts contain many important inter-sentence relationships, such as between code blocks and explanatory text, and tend to be much longer than 512 tokens (\Cref{fig:seqLength}). Toolkits for training LLMs now routinely use 2048 or more tokens for context windows.

We demonstrate in this work that the key to producing state-of-the-art models for (at least some) SE tasks lies in \emph{training software-specific models with LLM training practices}. We train two new BERT-style models on StackOverflow data: SOBertBase and SOBertLarge. We ground their training approach in key characteristics of the dataset and ML best practices. We first collect 19 GB of SO data, where each sample contains an entire post with its comments. We set the context window to a relatively high threshold of 2048 tokens and train a custom vocabulary to maximize the number of characters per token and thus the number of posts that are presented without truncation. In line with established scaling laws \cite{hoffmann2022training}, we train two relatively small models on this data for multiple epochs (29B tokens seen in total): SOBertBase has 125M parameters, and SOBertLarge has 762M parameters. We adopt hyper-parameters that ensure that both models are trained to high performance on four GPUs. Training these models would cost just ca. \added{\$374} and \added{\$1600} respectively with cloud resources.

We evaluate our models on four StackOverflow labeling tasks. On each task, we fine-tune both the SOBert models as well as three baseline BERT models, GPT-3.5 and GPT-4 prompted in multiple ways, and (for the final task) LoRA \cite{hu2021lora} fine-tuned LLaMa models with 7B parameters. We study four different tasks that measure SO question \& answer understanding: question quality prediction \cite{KaggleQuality}, closed question prediction \cite{KaggleClosedPred}, named entity recognition \cite{tabassum2020code} and obsoletion prediction. The latter is a new task that we introduce on detecting whether answers contain outdated code, a common phenomenon on SO that can cause vulnerability and compatibility issues for those adopting the code without scrutiny \cite{fischer2017stack,meng2018secure,ragkhitwetsagul2019toxic}. 
SOBert consistently, and often widely, outperforms all other compared models on all tasks, with even the smaller model, SOBertBase frequently yielding state-of-the-art performance.

These results highlight the significant promise of pre-training sub-billion parameter LMs on in-domain data when such data is rich in detail and abundant. We release our models for public use.
In summary we make the following contributions:
\begin{itemize}
    \item We advocate for a critical approach towards the adoption of LLMs on code-related tasks. Given their cost and proprietary nature, their performance should be judged against diligently trained open-source alternatives. 
    \item To demonstrate this, we introduce two models, SOBertBase (125M parameters) and SOBertLarge (762M parameters), that were trained on SO answer and comment text using the Megatron Toolkit. The training process used multiple GPUs and cost a modest budget of \added{\$374} and \added{\$1600} for SOBertBase and SOBertLarge, respectively.
    \item We extensively evaluate both SOBert models on four different SO tasks: question quality prediction, closed question prediction, named entity recognition, and obsoletion prediction. 
    \item Compared to three other variants of BERT, GPT-3.5, GPT-4, CodeLlama, StackLlama, and various fine-tuning and prompt engineering strategies, SOBert outperforms all models on all downstream tasks.
\end{itemize}

\begin{figure}[t]
    \centering
    \includegraphics[width=0.5\textwidth]{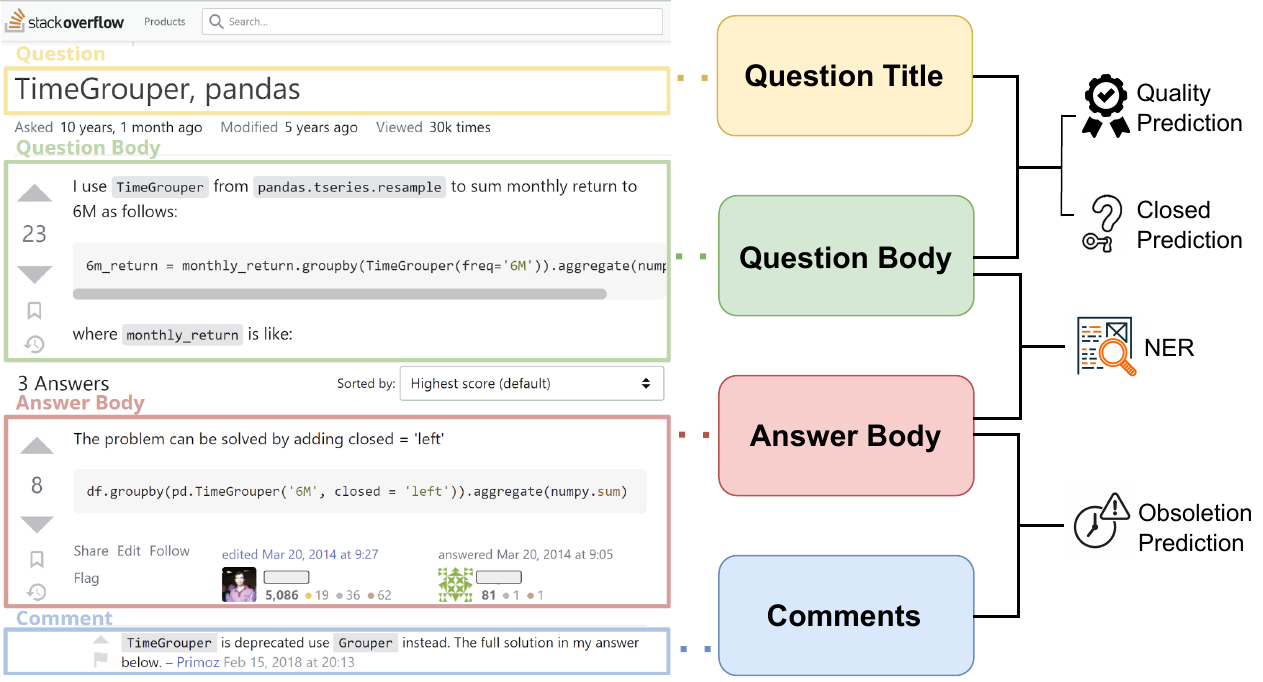}
    \caption{Example StackOverflow page (ID: 14569223) demonstrating the structure of a question and answer on SO. Questions have a title and body and may have comments; answers have a body and, optionally, comments. We demonstrate which features are used for each of the four downstream tasks.}
    \vspace{-15pt}
    \label{fig:Parts}
\end{figure}

\section{Background}
\label{sec:background}

To provide context and reduce ambiguity for the features discussed later, we begin by defining some of the terms we will be using in this paper. We then delve into the background work that has been done in this field.

Our case study, StackOverflow (hereafter abbreviated as SO), is a Q\&A platform. Pages on SO consist of the following main components: question title, question body, and or more answers each with a body and (often) comments. The title headlines and summarizes the post, the question body explains the issue, and the answer body provides a response. Comments offer additional remarks for clarification, feedback, or suggestions. Following the SO data terminology, a ``Post" is either a question with its title \& comments or an answer with its comments (so not the entire Q\&A page). Posts additionally come with metadata, such as tags that describe the topic, and votes that gauge post quality. Figure \ref{fig:Parts} depicts some of these elements in relation to our downstream tasks.

\begin{figure}
\includegraphics[width=0.48\textwidth]{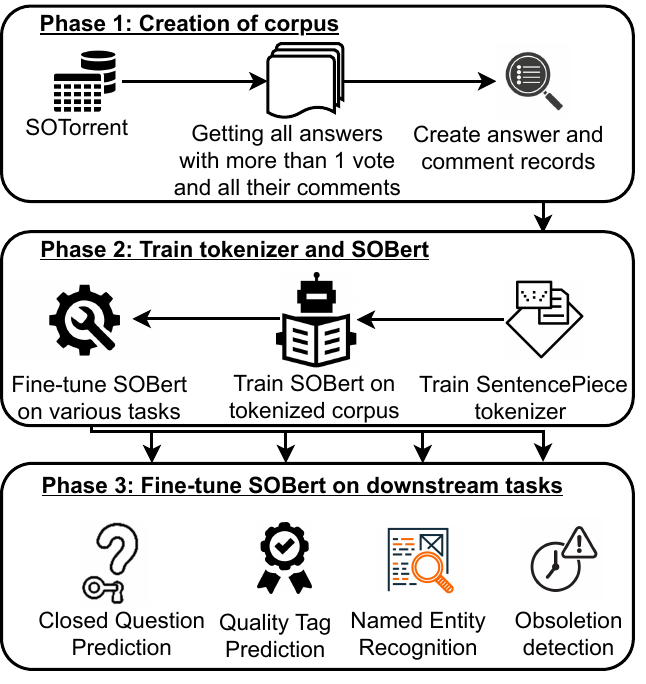}
\label{fig:Steps}
\caption{Framework outlining the key steps of our approach}
\vspace{-15pt}
\end{figure}

\subsection{Pre-trained Language Models}
Language models generate text in a given context. Auto-regressive, or left-to-right, language models do so by repeatedly predicting the next token in an input conditioned on all previous tokens, while masked-language models predict a random subset (typically some 15\%) of ``masked" tokens in an entire input. Both can be pre-trained on very large datasets of text by converting each document into a series of training examples. Auto-regressive, or decoder-only, models tend to be useful for downstream tasks where new text is a desired output, as in ChatGPT, while masked-language model training assumes that the entire input document (minus zero or more masked-out tokens) is available  while training which makes it easier to ``embed" inputs in a single representation that may then be used for learning any given downstream task.

BERT (Bidirectional Encoder Representations from Transformers) is a type of pre-trained masked language model developed by Google introduced in \cite{devlin2018bert} that can be fine-tuned for various natural language processing tasks. It utilizes the Transformer architecture \cite{vaswani2017attention}, which employs a self-attention mechanism to learn a contextual representation of text. It is trained using both the masked language modeling and next sentence prediction objectives on BooksCorpus (800M words) and Wikipedia (2,500M words). 
Of these two tasks, predicting masked words has proven particularly important as it teaches the model to represent each word in the context of its surrounding words, which helps it learn meaningful representations of the entire document as well. This makes BERT-like models versatile and effective in various downstream NLP tasks such as text classification \cite{sun2019fine,gonzalez2020comparing} and named entity recognition \cite{labusch2019bert,liu2022conll}.

We evaluate SOBert in comparison to seven pre-trained models. BERTBase-uncased \cite{devlin2018bert} is the smaller model, which has 12 layers of transformer blocks with an internal dimension of 768 for a total of 110 million parameters. The "uncased" tag means that its input text has been converted to lowercase and all punctuation has been removed, which simplifies learning from a heterogeneous body of text. BERTLarge \cite{devlin2018bert}, has 24 1,024-dimensional transformer layers for a total of 340 million parameters. We match our own Base model, SOBertBase, to the former, while adopting a model with a slightly larger hidden dimension (1,536) than BERTLarge but the same number of layers for SOBertLarge, yielding 762M parameters. We pre-train our models with the masked language modeling objective.

We also compare SOBert with LLMs from two model families for the obsoletion detection task. These include GPT-4 and GPT-3.5 as closed-source LLMs. We choose CodeLlama and StackLlama as open-source LLMs. 


\subsection{Modeling StackOverflow data} 
 SO frequently releases `data dumps' of publicly available content (questions, answers, comments, etc.). This, combined with its natural alignment of informative text and code, has made it a popular target for SE research. Numerous aspects of SE have been studied on SO by researchers. For instance, one study scrutinized the interactions among developers on the platform \cite{wang2013empirical}, while another analyzed the themes and patterns in discussions on SO \cite{barua2014developers}. Working with SO data is not without challenges. For one, the quality and structure of answers varies widely from post to post, with many answers being incomplete, containing errors, or lacking context. Another issue is the sheer volume of data on the platform, making it difficult to effectively navigate and identify high-quality task-relevant posts for specific study goals. To mitigate this, studies often focus on a narrow slice of the data, such as investigating the specific inquiries made by software engineers about Android testing on SO \cite{villanes2017software}.


\added{Post2Vec, as introduced in \citeauthor{xu2021post2vec}'s work \cite{xu2021post2vec}, represents one of the pioneering methods for exploring the generation of distributed representations for Stack Overflow posts. Post2Vec achieved this through the utilization of Convolutional Neural Networks (CNNs) as the primary feature extraction mechanism.} At the intersection of language models and SO data, there has been a lot of work in the space of fine-tuning BERT for various tasks such as emotion recognition \cite{bleyl2022emotion}, sentiment analysis \cite{biswas2020achieving}, automated summarization \cite{kou2023automated} etc. that involve SO data. Several models have been pre-trained on SO data specifically.  \citeauthor{he2022ptm4tag} \cite{he2022ptm4tag} build a tag recommendation framework PTM4TAG for Stack Overflow posts with different BERT-based models.  The model seBERT \cite{von2021validity} is a 340 million parameter model trained on 62.8GB of textual data (they remove all code from the posts) from answer, posts and comments from SO along with more data from GitHub and Jira using the BERTLarge architecture. They use a case-insensitive 30,522 token WordPiece vocabulary to tokenize their data and restrict inputs to a maximum length of 128 tokens -- significantly shorter than most SO posts (Figure \ref{fig:seqLength}). \citeauthor{tabassum2020code}\cite{tabassum2020code} released a BERT model pre-trained on SO data. Their model, BERTOverflow, was pre-trained on 152 million sentences from SO using a case-sensitive 64,000 token WordPiece vocabulary. Training on individual sentences was a common practice in training BERT-like models, but recent work has found that language models perform substantially better when considering a document-level context \cite{kaplan2020scaling}, albeit at a substantial increase in compute cost. Our findings echo that observation. \citeauthor{tabassum2020code} focus on the downstream task of Named Entity Recognition (NER, essentially tagging each word with its syntactic role), on which their model out-performs existing models such as BERTLarge as well as GloVe and ELMo models trained on the same SO data.

\textbf{Selection Rationale for Comparative Models: } We selected BertOverflow as the main specialized baseline in our work. We did not include comparisons to the other models due to differences in model architecture and training data between these models and our proposed SOBert. Specifically, seBERT incorporates additional data sources beyond SO during pre-training, while PTM4Tag only models the Title, Description, and Code from SO posts. We omitted a comparison with Post2Vec because its performance has been substantially surpassed by Transformer-based approaches, e.g., as shown in PTM4Tag \cite{he2022ptm4tag} on a tag prediction task. Instead, we include BERT-Base and BERT-Large as strong, general models, in line with our broader goal of demonstrating that domain-specific pre-trained models can outperform general models. And we include BertOverflow as a representative, capable domain-specific model to highlight how adopting LLM training practices allows for training significantly more powerful domain-specific models.

We selected GPT-3.5 and GPT-4 for comparison due to their strong general language understanding capabilities and widespread adoption as closed-source LLMs. We also considered the open-source CodeLLaMA-Instruct \cite{codellama} and StackLLaMA \cite{stackLlama} models, which are pre-trained on code and natural language data, respectively, to leverage their domain-specific knowledge and capabilities. We used the instruction-tuned variant of CodeLLaMA, while StackLLaMA is trained on pairs of questions and answers from Stack Exchange using Direct Preference Optimization (DPO) \cite{dpo} with the Transformer Reinforcement Learning (TRL) library \cite{trl}.

Compared to these efforts, we pre-train SOBert with 19 GB data presented as just 15 million samples where each sample contains an entire post and all its corresponding comments. We also include all code in each answer so that our model is bimodal in nature. Moreover, we use a SentencePiece tokenizer trained with Byte-Pair Encoding \cite{kudo2018sentencepiece}, which has the benefit over WordPiece of never labeling tokens as ``unknown". Additionally, SOBert is trained with a maximum sequence length of 2048, informed by the empirical length distribution of SO posts (see Figure \ref{fig:seqLength}) and a relatively large batch size of 0.5M tokens (following \citeauthor{kaplan2020scaling} \cite{kaplan2020scaling}). Our model performs substantially better than both BertOverflow and BERT models trained on general corpora of text on all downstream tasks (see Section \ref{sec:discussion}). 


\vspace{-5pt}
\section{Approach}
 SO contains code in various languages with rich natural language information about the code. In this section we describe how we design SOBert, a BERT model trained on bimodal data from 15 million SO questions and answers.

\subsection{Data and Preprocessing}
We downloaded the SO data dump published on 7 March 2022 \cite{SO} which includes posts from 2008-2022.
The SO data dump is a collection of structured data that includes all information publicly available on the website. The data is made available for download periodically and includes information such as user profiles, questions, answers, comments, tags, and votes. The data is provided in XML format and is compressed into a set of files that can be downloaded and processed using a variety of tools and programming languages. Each file corresponds to a specific type of data, and contains all the relevant information for that type. We loaded these files into SQL database tables and specifically worked with the `Posts' and `Comments' tables. 
\added{Using the entire corpus of SO posts would have introduced a significant amount of low quality, unengaged content that may not be representative of posts software engineers typically rely on. We thus make a design choice to prioritize quality content that demonstrated some level of community engagement by filtering the answer posts to extract only those that meet the criteria of having a minimum of one up-vote and at least one accompanying comment. 
We then use this filter and extract both answer posts along with all their associated comments.} This yielded 15 million answers and 39.5 million comments (median 2, mean 2.68 comments per post).

\vspace{-5pt}
\begin{figure}
  \begin{center}
    \includegraphics[width=0.5\textwidth]{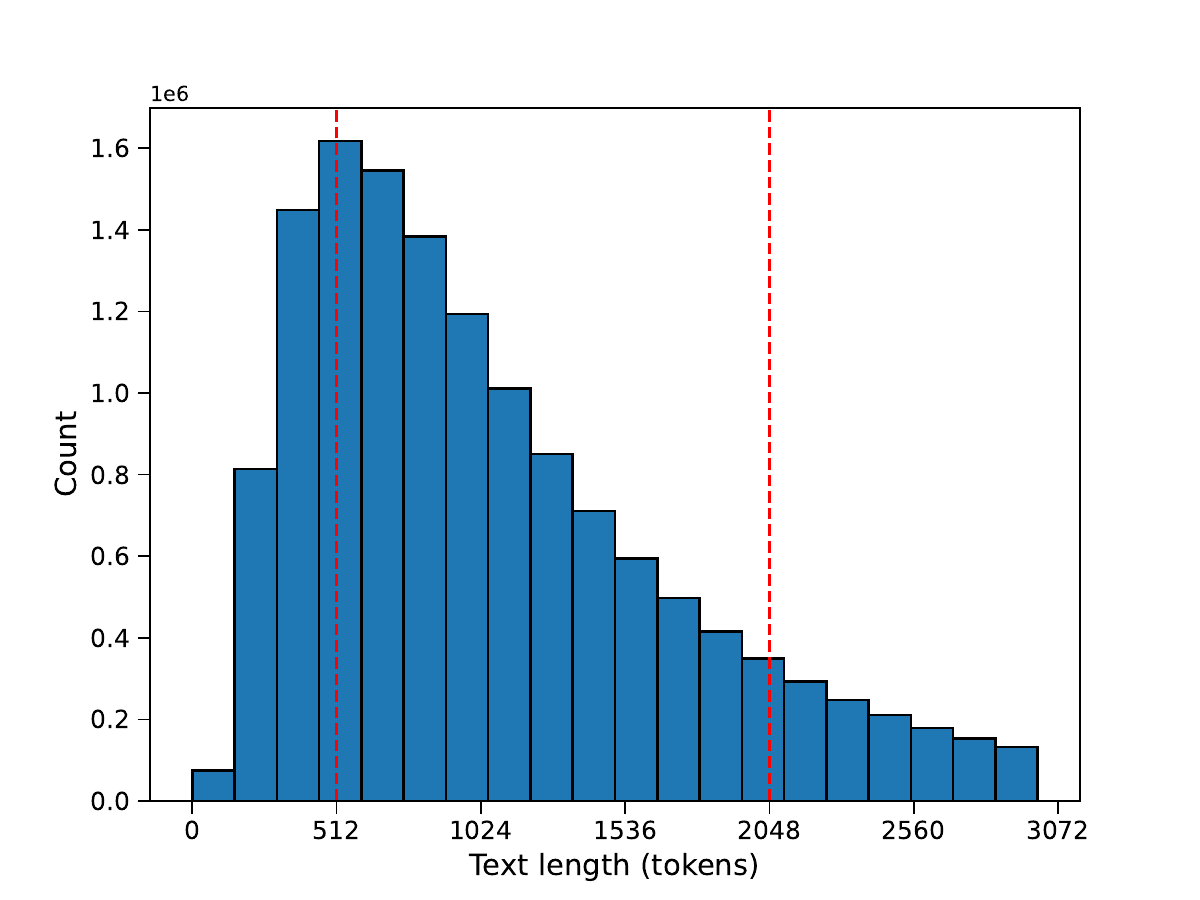}
    
    \caption{Histogram showing length buckets of post+comments samples}
    \label{fig:seqLength}
    \vspace{-25pt}
  \end{center}
\end{figure} 

\subsection{Data decisions}

\citeauthor{devlin2018bert} \cite{devlin2018bert} and \citeauthor{tabassum2020code} \cite{tabassum2020code} tokenize the datasets for BERT and BERTOverflow respectively using a WordPiece vocabulary with 30,522 tokens. We preprocess the data for SOBert using a 50,000 token Byte-Pair Encoded (BPE) vocabulary, trained using SentencePiece \cite{kudo2018sentencepiece}. BPE sub-tokenizes words by greedily finding the largest tokens in its vocabulary that collectively reconstruct the word. While common words such as ``the" tend to appear as a single token in the vocabulary, rare words may be broken up into smaller pieces. Since BPE uses a large vocabulary of single characters as its base vocabulary, it avoids treating almost any token as ``unknown" or out-of-vocabulary, in contrast to WordPiece, which tends to improve results on downstream tasks \cite{karampatsis2020big}. We trained our tokenizer using a randomly selected subset of 10\% of the entire dataset (ca. 2GB) due to memory limitations. The input text is first cleaned \added{ following standard data cleaning procedures such as first replacing URLs and email addresses with generic [URL] and [EMAIL] tokens to abstract away identifiable information.}

We use Beautiful Soup \cite{beautifulsoup} to remove all HTML tags except for \texttt{<code>} tags. We make sure to keep blocks of code, demarcated with \texttt{<code>} \texttt{</code>} tags, in the input data to preserve the complete structure of each post. This ensures that any code snippets or programming language syntax are not altered or misrepresented during processing or analysis. The preprocessed text is then input into the SentencePiece tokenizer for subword tokenization. A reserved token, \texttt{<RS>}, is placed between the answer and each of the comments to signal to the model where each comment begins. This tokenized representation is then passed through a stack of Transformer layers, described next.

\subsection{Model Design}
\label{sec:design}

SOBert was pre-trained using the Masked Language Model (MLM) objective. This involves predicting masked tokens within a sentence, helping the model to learn the statistical properties of the language. The MLM objective can be represented by the following formula:
\vspace{-7pt}
\[
\text{MLM Loss} = - \sum_{i=1}^{N} \log P(w_i | \text{context})
\]

Where \( N \) represents the total number of masked tokens, \( w_i \) represents each masked token, and \( P(w_i | \text{context}) \) represents the probability of predicting the correct token given the surrounding context.

For the BERT base model the maximum length of the input sequence for the model is 512 tokens and each token is represented with a 768 dimensional vector. The choice of 2048 as the maximum length of the input sequence for SOBert was informed by an analysis of the text lengths of combined posts + comments that we use as inputs to our approach, shown in Figure \ref{fig:seqLength}. This shows that rather few posts are no more than 512 tokens long, which has conventionally been used as the maximum input length \cite{feng2020codebert,wang2021codet5}. A slight majority of the posts contain up to 1024 tokens but many more posts span between 1024-2048 tokens. As the cost of encoding long inputs grows quadratically in attention-based models, we choose to set the maximum length of inputs in our study to 2048 to balance training cost and fidelity to the task domain.

\begin{figure}
  \begin{center}
    \includegraphics[trim=0 0 0 1.25cm,width=0.45\textwidth]{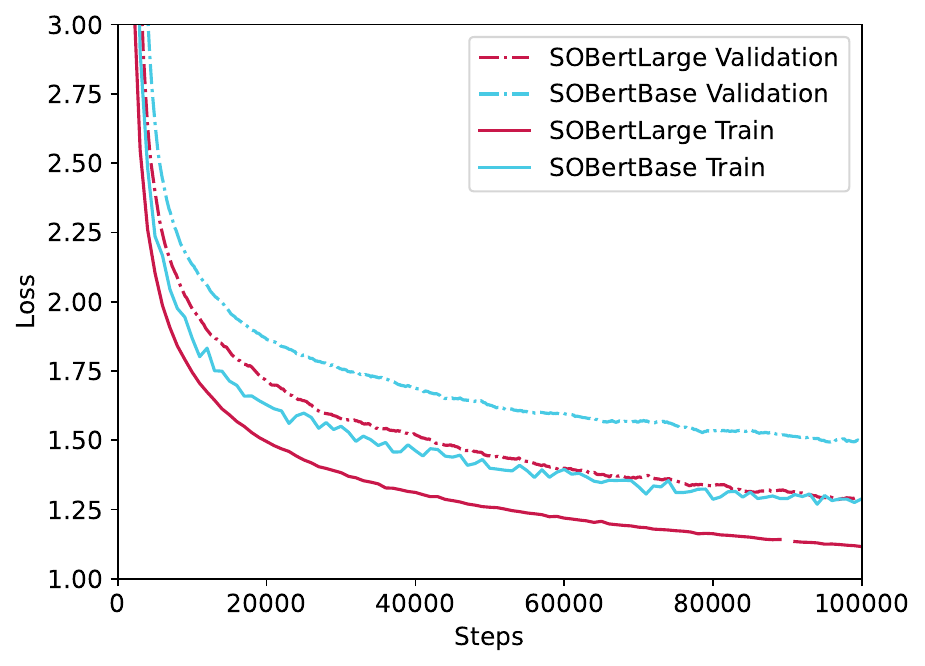}
    \caption{Training and Validation Loss comparison of SOBBase and SOBLarge.}
    \label{fig:lossPlotsComparison}
    \vspace{-27pt}
  \end{center}
\end{figure}

 
Batch size refers to the number of training examples that are processed together in a single iteration during training. Larger batch sizes tend to lead to better language model performance 
\cite{kaplan2020scaling} by providing more stable gradients for each model update, but they also require more memory and computing resources. We use gradient accumulation to simulate large batch sizes on our limited hardware, allowing us to process the training data in batches of 0.5 million tokens for both models. DeepMind proposed a series of ``scaling laws" that can be used to determine how to best balance a given compute budget between training larger models and providing models more data \cite{hoffmann2022training}. Their conclusion was that models should be trained with ca. 20 times more tokens than they have parameters. While models trained this way won't have converged (training longer would continue to yield better performance), the empirical observation is that this minimizes loss for a given compute budget. For instance, if one had the budget to train a 1 billion parameter model with 50B tokens, it would instead be better to train a significantly larger (e.g., 1.5B parameter) model with 30B tokens. The latter would achieve a lower loss in the same training time. This ``rule" thus provides a useful lower bound for the training budget that should be allocated to a given model. We determined that, with our compute budget (see Section \ref{sec:details}), the largest model that we would be able to train to this ``optimum" within ca. two weeks time contained ca. 760M parameters. We therefore aimed to train this model with at least 15B tokens. Our actual training resources enabled us to provide nearly twice as many tokens as this minimum, which naturally improved its performance (Figure \ref{fig:lossPlotsComparison}).\footnote{We follow the conventional GPT-2 architecture sizes here. The typical ``next-size" model would contain 1.4B parameters, which we estimate we could only train with 10 - 15B tokens, falling well short of the compute-optimum.}

We train our models using the Megatron-LM toolkit \cite{shoeybi2019megatron}. This toolkit was introduced for efficient parallel model training of LLMs with up to 8.3B parameters. It provides powerful support for multi-GPU training with optimized kernels for many components. It also incorporates many architectural best-practices.  \citeauthor{shoeybi2019megatron}\cite{shoeybi2019megatron} showed that rearranging the order of the layer normalization and the residual connections is critical to enabling the scaling of the BERT-style models beyond 336M parameters. We use the same architecture configuration for SOBert.

\subsection{Training details}
\label{sec:details}
The training was performed on four NVIDIA Quadro RTX 8000 GPUs, each with 48GB of memory, hosted in an on-premise server. To calculate the cost of training our models for those without access to this hardware, we base our estimates on quotes for NVIDIA's A6000 GPU series, as these are currently widely available via cloud computing environments and have the same memory capacity as our GPUs. A6000 GPUs outperform the Quadro RTX 8000 by ca. 80\% \cite{gpuPerformance}, so we take the GPU hours required to train each of our models and multiply them by the standard rate of an A6000 in the cloud, which is approximately \$1 per hour \cite{gpuPricing}, and by the ratio 1/1.8 to compensate for the performance disparity. Training the SOBertLarge model required ca. \added{30} days on four GPUs for a total of \added{ca. 2,880} GPU hours at an A6000-equivalent cost of \added{\$1600}. The SOBertBase model required just \added{7} days (\added{ca. 672} GPU hours) to train, which amounts to \added{\$374}.

Figure \ref{fig:lossPlotsComparison} shows the comparison between the training and validation loss curves of SOBertBase and SOBertLarge. \added{Both SOBertBase and SOBertLarge were trained for 100,000 steps at which point they had seen around 54 billion tokens.}

\subsection{Limitations}

While our assessment of SOBert includes a comprehensive analysis of four downstream tasks from well-established sources, it is important to recognize that the applicability of our findings to different tasks may differ. Given that SOBert is a predictive model, our focus is primarily on classification tasks rather than generative ones. Although we would have preferred to test baselines against multiple BERT models pre-trained on SO data such as SeBERT and PTM4Tag, resource limitations compelled us to select only the state-of-the-art BERTOverflow and two general BERT models of varying sizes. We discuss our selection rationale in detail in Section \ref{sec:background}. Additionally, the nature of API access from OpenAI implies that the performance of the language model can fluctuate over time due to model updates without prior notification, potentially affecting its performance. However, the fact that the LLMs are much less accurate in each task implies that it is unlikely to be able to bridge the absence of task specific training with examples alone.

\section{Evaluation}
\label{sec:evaluation}
In this section, we establish the metrics employed for evaluating and contrasting the models, while providing a comprehensive overview of the four SO related downstream tasks. SO is known to cover numerous SE topics and attracts numerous software developers. Among these, three tasks have been sourced from prior research or Kaggle competitions, namely question quality prediction \cite{KaggleQuality}, closed question prediction \cite{KaggleClosedPred}, and named entity recognition (NER) \cite{tabassum2020code}. Additionally, we introduce a novel task we have formulated, known as obsoletion detection. We try to cover different types of downstream tasks with varying amounts of data available for fine-tuning. The above tasks, described in detail below, include a multi-class classification problem with about 60k samples, a multi-class classification problem with 140k samples, a token-level tagging task with 15k samples. The new task, obsoletion detection, is a binary class classification task with just 942 samples.

\textbf{Selection Rationale for downstream tasks:} Despite some claims that SO is losing relevance, the demand for human-generated responses to novel programming inquiries will persist and cannot be entirely supplanted by current technological advancements \cite{kabir2023answers}. We have selected these four tasks based on their significance within the SE community, their relevance to the SO data source we used for training, and the availability of data for effective model performance comparisons. The first task involves predicting the quality of questions on platforms like SO. By identifying low-quality questions, this task contributes to a more efficient and informative knowledge-sharing environment. The second task predicts question closures, assisting moderators in managing platform content and ensuring its quality. NER, the first task, extracts crucial technical entities from text, benefiting various SE tasks such as code search by linking APIs and libraries to documentation. Finally, obsoletion detection, our fourth task, plays a vital role in highlighting outdated code snippets, promoting code health and security by preventing the use of vulnerable, deprecated code. These tasks are also relevant for evaluating the quality and safety of code generated by LLMs, which are increasingly being adopted by the SE community. These tasks collectively empower the development of improved tools for software developers, ultimately contributing to a more efficient and secure SE environment.

\textbf{Fine-tuning details}: Our evaluation of the models involves a comparative analysis with two general-purpose BERT models, namely BERTBase \cite{devlin2018bert} and BERTLarge \cite{devlin2018bert}. In addition, we also assess the performance of the models against a domain-specific model called BERTOverflow \cite{tabassum2020code}. In all cases, each model is trained with the same hyper-parameters, typically involving a batch size of 32 samples and a learning rate of 1e-5. Small variations from task to task, mainly in the input length, are noted below. We note that, although our model is capable of handling longer sequences, we set the maximum sequence length to be the same as that of the baseline models for all our experiments. 

We also compare SOBert with LLMs from two model families for the obsoletion detection task. These include GPT-4 and GPT-3.5 as closed-source LLMs. We choose CodeLlama and StackLlama as open-source LLMs. We use zero-shot, few shot with CoT prompting for the closed-source LLMs.  We fine-tuned the open-source LLMs using Quantized LoRA (QLoRA) and the Parameter-Efficient Fine-Tuning (PEFT) library. QLoRA involves training only a small set of quantized adapter weights instead of the full model weights, significantly reducing memory and computational requirements. We utilized the LoraConfig from the PEFT library to specify the QLoRA configuration and applied techniques like gradient checkpointing and mixed-precision training for efficient fine-tuning. The fine-tuning process was carried out using the SFTTrainer from the Transformers library. After fine-tuning, we merged the trained QLoRA adapter weights with the original CodeLLaMA model weights using PEFT's {\textit{merge\_and\_unload}} method, creating a single model file for inference. This approach enabled efficient adaptation of the large CodeLLaMA model while preserving its pre-trained knowledge.

\textbf{{Data leakage:}} During the pre-training phase, SOBert was exposed to a wide range of text from SO, which may or may not include some answers that appear in the downstream tasks. However, it did not have access to the labels for any of the downstream tasks. This reduces the risk of confounding effects related to data leakage in downstream tasks.

\subsection{Task 1: {\includegraphics[height=1em]{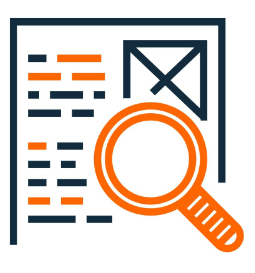}} Named entity recognition (NER) \cite{tabassum2020code}}

\citeauthor{tabassum2020code}\cite{tabassum2020code} release a dataset for the extraction of software-related named entities from StackOverflow. It spans roughly 10 years (from September 2008 to March 2018) of question-answer threads in which each token in each sentence is manually annotated with close to 50 types of entities. The StackOverflow NER corpus has 9,352 train, 2,942 development and 3,115 test sentences. We fine-tuned all the models with a batch size of 16, maximum sequence length of 128 (based on the maximum length of the samples in the dataset) and learning rate of 1e-4.\footnote{Hyper-parameters are particularly different here because this task involves one label per token, so a batch of 16 samples has far more labels than in the other, sequence-level prediction tasks.} We use a weighted cross entropy loss function as the dataset is imbalanced with many more `O' (no label) tags than others.

We were able to replicate the results of \cite{tabassum2020code} within a 5\%-10\% margin for all reported metrics with this configuration on both their dev and test sets. The results of this task are discussed in the next section.

\subsection{Task 2: {\includegraphics[height=1em]{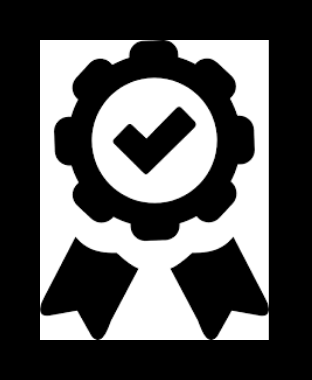}} Quality Tag prediction \cite{KaggleQuality}} 
We utilize a Kaggle competition dataset focused on predicting question quality on SO for content moderation in our downstream task \cite{KaggleQuality}. This dataset features 60,000 SO questions from 2016-2020 classified into three categories based on their quality - HQ: High-quality posts without a single edit, LQ\_EDIT: Low-quality posts with a negative score, and multiple community edits. However, they still remain open after those changes, LQ\_CLOSE: Low-quality posts that were closed by the community without a single edit.

We use a batch size of 32, learning rate of 1e-5 and set the maximum sequence length as 512 for this task. The results of this task are discussed in the next section.

\subsection{Task 3: {\includegraphics[height=1em]{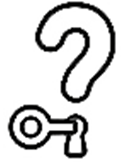}}Closed question prediction \cite{KaggleClosedPred}}
We choose a Kaggle competition dataset for predicting closed questions on SO \cite{KaggleClosedPred} for this downstream task. The provided dataset contains 140,272 samples. The competition motivates this task by saying that SO is a widely-used platform that provides programmers with high-quality answers to their programming questions on a daily basis. At the time this competition was published, SO received over 6,000 questions every weekday and an automatic content moderation would help maintain post quality. One aspect of moderation involves closing questions that are deemed inappropriate for the platform. At the time this competition was published, around 6\% of all new questions ended up being closed for various reasons. The purpose of the competition is to analyze said quality and predict when questions should be closed for effective moderation. They excluded the `exact duplicate' closed reason from the dataset as it depends on previous questions. The competition requires the development of a classifier that can predict whether a question will be closed or not, based on the submitted question and the reason for closure. The categories of classification are `open', `not a real question', `off topic', `not constructive' and `too localized'.

The competition provides a plethora of features from user data, however for simplicity we use the question title and question body for the classification task at hand. We use a batch size of 32, learning rate of 1e-5 and set the maximum sequence length as 512 for this task. The results of this task are discussed in the next section.

\subsection{Task 4: {\includegraphics[height=1em]{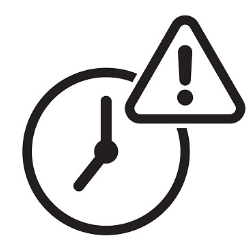}}Obsolete post detection {\includegraphics[height=1.25em]{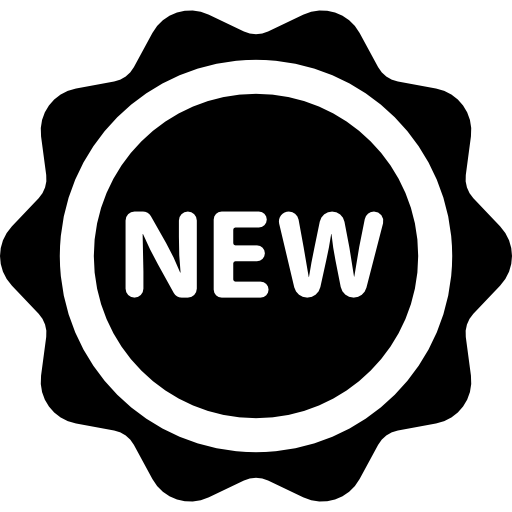}}}

\begin{figure}
  \begin{center}
    \includegraphics[width=0.45\textwidth]{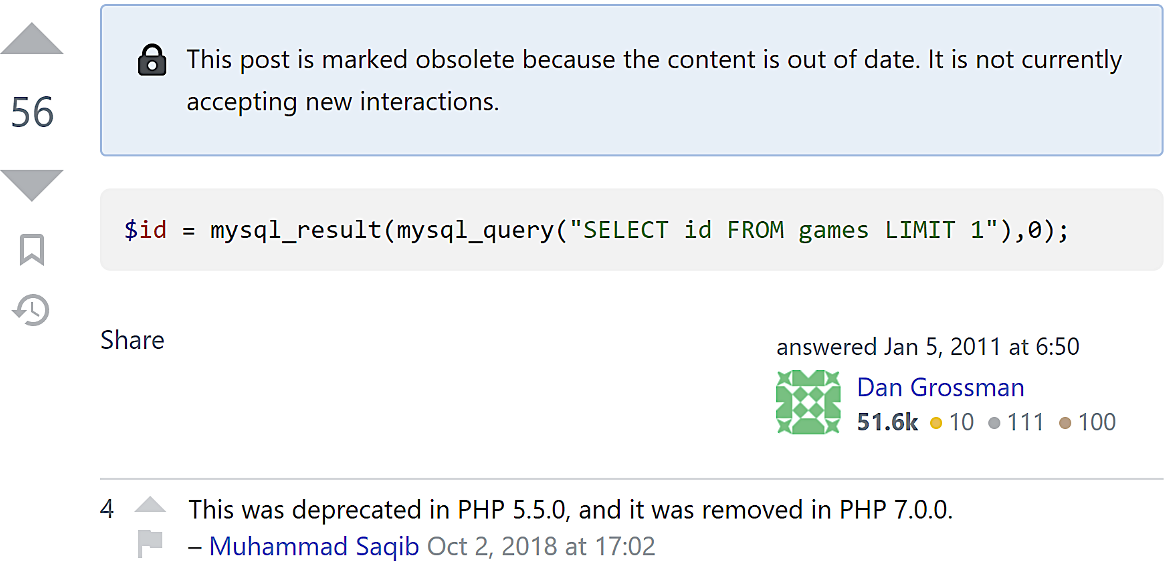}
  \end{center}
  \caption{Answer 4601538 marked as obsolete on StackOverflow}
  \vspace{-15pt}
   \label{fig:obsoleteAns}
\end{figure}

Given the speed at which technology advances, answers on StackOverflow often become outdated. Outdated answers are any of those that propose solutions that are no longer state-of-the-art. They may offer solutions that are less performant or precise than what is now available, use now-deprecated APIs and libraries (be obsolete), and in some cases propose solutions that create security concerns. For instance, there have been more than 25 versions of TensorFlow \cite{tf} since its release in 2015, including a major change to its entire compute structure (version 2). This library is also a highly popular subject of questions. Many of those are now years old and involve classes, methods, and parameters that have since been renamed. More generally, over 24 million answers contain code snippets, which software engineers often hope to reuse. When such snippets are obsolete, that reuse may severely reduce the quality of their code.
One might hope that such ``outdatedness" is easily discovered by e.g. looking at the date posted, or comments pointing out such issues. However, date is not always a good indicator of obsolence (one type of outdatedness); \citeauthor{zhang2019empirical} \cite{zhang2019empirical} found that more than half of obsolete answers were already obsolete at the time of posting. Indeed, `year of posting' only poorly explained outdatedness in our analysis as well. Comments are a useful resource: many posts have comments to indicate obsoletion. However, commenting or editing answers is time-consuming, so many posts with obsolete content go unmarked. Downvoting obsolete answers is ineffective, as posts that were once correct and popular require many downvotes to yield a negative score. SO’s maintainers are aware of this problem and have recently proposed explicitly marking outdated answers as such as shown in Figure \ref{fig:obsoleteAns}. However, there are only 9 such answers marked as obsolete by SO at the time this paper was written. We thus propose to fine-tune SOBert to be able to predict these obsolete answers based on a manually curated dataset, which we describe next.

\subsubsection{Raw Data Collection} 
We curate an initial dataset of obsolete answer candidates based on three heuristics described below, which are selected to provide the model with a wide range of patterns to learn from. In all cases, we focused only on answers with at least some community support, filtering out answers which had less than 1 upvote to reduce noise from potentially bad solutions.

\paragraph{Comments With Keywords}
\label{sec:comments}
Users often leave comments on a post or answer to indicate obsoletion. We selected answer threads on the basis of the selection criteria from \cite{zhang2019empirical}. Comments were chosen which had one of the four keywords: \textit{deprecated}, \textit{outdated}, \textit{obsolete} or \textit{out of date}, while the corresponding question posts did not have these keywords in them. We found 85,586 such samples of comments.

\paragraph{Answers Edited After Comment}
Often, though far from universally, when comments point out issues with an existing answer, that answer is edited to resolve that concern. Yet it is also not uncommon for answers to be edited even without such prompting, sometimes for cosmetic reasons, often soon after posting. Other edits are more substantial and often dramatically revise and extend an answer. To prioritize discovering the latter kind, we chose answers that were updated after a comment was posted and the Levenshtein distance between the code before and after edits was at least 100, indicating a significant change to the answer. We also filter out answers which have been included while considering the comments with keywords to prevent these answers from being repeated in our dataset. We find 388,809 samples of answers edited substantially following a comment.

\paragraph{Answers Added Late}
\label{sec:late_answers}
Often, questions continue to draw attention for years after posting from developers facing similar issues, but the originally accepted answer ceases to be relevant. When this happens, the simple solution would be to simply edit the existing answer, but for various reasons\footnote{Some related to SO's ``reputation" system, which prevents all-but highly experienced users from editing others' answers.} this is not always the path taken. Rather, we found many cases where a new answer was posted long after the original/accepted one, which usually references the previous answers in some way.
These may serve various purposes, such as showing how a new feature can be used for a better solution than what was previously posted.
We included such answers that were posted 1.5 years or more after the initial question was posted, where the answer had more than one upvote and referred to another answer on the same question thread. The check for reference to another answer is heuristic and keyword based, where we look for the keywords \textit{'s answers}, \textit{answer by}, \textit{accepted answer} and \textit{other answer}. We found 19,371 such samples.

\subsubsection{Data Annotation}
The goal of our annotation process is to produce samples consisting solely of the text in an answer or comment on the answer along with a classification of obsolete or otherwise. We first apply our heuristic selection criteria described above to obtain some 493K samples from the total database. We manually annotate 1000 random samples out of these, sampling equally from the three categories. Answers or comments that either include or refer to obsolete code in any way (e.g., comments pointing out obsolete code in the answer or answers edited to point out they contain outdated solutions) are classified as obsolete, regardless of whether newer techniques or alternative solutions are also provided in the same. This way, models trained on this dataset can be used to highlight posts that should be marked as obsolete using SO's new system based on existing community input.

The annotation protocol was developed by having two professional programmers annotate a pilot set of answers independently. After completion of the pilot annotation, the differences and difficult-to-annotate cases were discussed and details of such cases were documented. This protocol was improved iteratively by both professionals on other pilot sample sets until consensus was reached about the protocol. The data annotation process relied on the protocol and a set of 10 annotated samples that served as guidelines. The pilot annotation sets were then discarded and the same samples were re-annotated using the finalized protocol. The inter-annotator agreement of the annotations using the finalized protocol has a Cohen’s kappa of 0.63, which indicates that the agreement level is substantial \cite{landis1977application}.

This yielded many non-obsolete examples -- while many comments and new answers contain the keywords we relied on, and many answers are heavily edited over time, these often signified spurious concerns, so the task for the models is nontrivial.
We fine-tuned all the models with a batch size of 32, maximum sequence length of 512 (this being the largest sequence length supported by the baseline models) using a learning rate of 1e-4. We use a weighted cross entropy loss function as the dataset is imbalanced with many more non-obsolete samples than obsolete ones.

\section{Results}
\label{sec:discussion}

In the NER task, we utilized the test set provided by the authors and observed a significant improvement of +22 points in the F1-score when comparing SOBertBase to BERTOverflow as shown in Table \ref{tab:results}. We did not compute the balanced accuracy of models other than the SOBert models, since the original paper \cite{tabassum2020code} did not incorporate it in their evaluation metrics. When compared to the values reported in the paper for SoftNER, which incorporates a context-independent code token classifier with corpus-level features on top of BERTOverflow to further increase performance, we found that both SOBert models had an improvement of ca. +11 (SOBertLarge) and +4 (SOBertBase) in F1-score over SoftNER.

In the second task quality tag prediction, we found the SOBertLarge model to be particularly powerful, yielding a +20 point F1-score improvement compared to BERTOverflow. Our smaller, SOBertBase also improved over BERTOverflow's F1 score by +14 points and by +15 points over BERTLarge.

\begin{table}[t]

\begin{tabular}{p{1.3cm} p{1.4cm} p{1.3cm} p{1.3cm} p{0.5cm} p{0.5cm}}
\toprule
\textbf{Task} & \textbf{Model} & \textbf{Balanced Accuracy} & \textbf{Weighted Precision} & \textbf{Recall} & \textbf{F1-Score} \\
\midrule
\multirow{5}{*}{\includegraphics[height=1.5em]{sections/Images/3.png} NER} & BertOverflow & -- & 0.69 & 0.67 & 0.68
 \\
& BBU & -- & 0.80 & 0.46 & 0.55 \\
& BLU & 0.56 & 0.83 & 0.56 & 0.63
\\
& SOBertBase & \textbf{0.88} & \textbf{0.93} & 0.88 & \textbf{0.90} \\
& SOBertLarge & 0.83 & 0.86&\textbf{0.93}&0.83\\
& SoftNER & -- & 0.78 & 0.80 & 0.79
\\
\midrule
\multirow{5}{*}{\includegraphics[height=1.5em]{sections/Images/1.png} Quality} & BertOverflow & 0.72 & 0.72 & 0.72 & 0.72 \\
& BBU & 0.82 & 0.83 & 0.82 & 0.82 \\
& BLU & 0.71 & 0.72 & 0.71 & 0.71 \\

& GPT3.5@0 & 0.42 & 0.41 & 0.42 & 0.40 \\
& GPT4@0 & 0.54 & 0.53 & 0.54 & 0.47 \\
& GPT3.5@10 & 0.39 & 0.39 & 0.39 & 0.38 \\
& GPT4@10 & 0.46 & 0.35 & 0.34 & 0.30 \\

& SOBertBase & 0.86 & 0.86 & 0.86 & 0.86 \\
& SOBertLarge &\textbf{0.92} & \textbf{0.92} & \textbf{0.92} & \textbf{0.92} \\

\midrule
\multirow{5}{*}{\includegraphics[height=1.5em]{sections/Images/2.png} \shortstack{ Closed\\ Question}} & BertOverflow & 0.53 & 0.49 & 0.53 & 0.41 \\
& BBU & 0.60 & 0.56 & 0.60 & 0.56 \\
& BLU & 0.53 & 0.48 & 0.53 & 0.43 \\
& GPT3.5@0 & 0.23 & 0.23 & 0.23 & 0.16 \\
& GPT4@0 & 0.35 & 0.35 & 0.35 & 0.29 \\
& GPT3.5@10 & 0.30 & 0.25 & 0.30 & 0.24 \\
& GPT4@10 & 0.41 & 0.37 & 0.41 & 0.37 \\
& SOBertBase & \textbf{0.63 }& \textbf{0.62 }& \textbf{0.63 }& \textbf{0.61} \\
& SOBertLarge & \textbf{0.63} & 0.61 & \textbf{0.63} & \textbf{0.61} \\

\bottomrule
\end{tabular}
\caption{Performance Metrics for the Models Across the 3 Tasks - Named Entity Recognition (NER), Quality Prediction, and Closed Question Prediction }
\vspace{-20pt}
\label{tab:results}
\end{table}

In the third task, which involves closed question prediction, we observed an increase of +20 in F1-score between both  SOBert models and BERTOverflow. This marks the second task, after NER where our Large model is slightly outperformed by the smaller one -- we discuss this phenomenon below.  The F1-score for closed question prediction was relatively low for all the models in this task, with the highest F1-score achieved by SOBertBase and SOBertLarge at 0.61. While low, these scores are similar to the results reported in recent studies on the same task \cite{roy2020predicting}. The dataset used in this task contains several features, such as \textit{CreationDate, Tags, Reputation, Title, BodyMarkdown,} etc. \citeauthor{roy2020predicting}\cite{roy2020predicting} have explored the use of feature selection techniques and chose to use \textit{BodyMarkdown}. We used both the \textit{Title} and \textit{BodyMarkdown}.

In the fourth task (results in Table \ref{tab:ComparisonChatGpt}) which involves obsoletion detection, we see large gains on the F1-score between SOBert and all the other baselines models. While StackLlama's precision edges out SOBert by a slight margin, it offers substantially lower recall. There is also a substantial difference in their sizes, with StackLlama at 7B parameters being 10 times the size of SOBertLarge. We delve into more details in the following section.

\section{Discussion}

We started this work by arguing that, while general purpose LLMs perform well on many tasks, they do not guarantee cutting-edge performance for tasks where rich domain-specific data is available, but that this requires careful training of the domain-specific models. The results show that even our smallest model, SOBertBase, concistently outperforms both ChatGPT and several general-purpose BERT models. Our substantial improvement over BertOverflow further highlights that it is crucial to carefully evaluate the requirements, constraints and data available for a given problem when building models in data-constrained domains.

\subsection{Relation to BERT}
\label{ssec:bertOverflow}
In multiple experiments, the generic BERT models outperform BERT-Overflow despite the latter using a very similar architecture and training on StackOverflow data only. In two cases, quality tag and closed question prediction, they even came close to rivaling one of our SOBert models. Their strong performance can likely be attributed to a combination of two sources. For one, compared to BertOverflow, these BERT models were trained on significantly more data (16GB vs. 11GB), which would have helped them learn the patterns in language more effectively. For another, the two downstream datasets where they perform best are relatively large, containing 60K and 140K samples respectively. Fine-tuning on a larger dataset likely allowed these models to overcome much of the distribution-shift relative to their pre-training setup.

For the final task of obsoletion detection, we find the F1-score increased by a large margin between both SOBert models and BERTOverflow. Although BERTOverflow converges the fastest during fine-tuning between all BERT models we compared for this task, it struggles with this classification task. The structure of its training data is likely a core factor in its performance. During pretraining, it processed SO data as individual sentences in isolation. That removes much of the valuable context from their original SO post. Evidently, the representations learned this way depress performance even after fine-tuning with complete posts.

\subsection{Relation to model size} In this study, we noticed that both SOBert sizes perform better than both BERT sizes in all of the four tasks we examined. However, we did not find a consistent trend of the larger models (of either kind) outperforming their smaller counterparts. For example, in closed question prediction and obsoletion detection, SOBertBase outperformed SOBertLarge in terms of F1-score. Similarly, BLU did not perform as well as BBU in quality tag prediction. Out of the four tasks, the two large models both performed better in exactly half.
\begin{table}[ht]

\centering
\renewcommand{\arraystretch}{1.2}
\begin{tabularx}{\columnwidth}{@{}l>{\raggedright\arraybackslash}p{1.5cm} *{3}{>{\Centering\arraybackslash}X}@{}}
\toprule
\textbf{Method} & \textbf{Balanced Accuracy} & \textbf{Precision} & \textbf{Recall} & \textbf{F1} \\
\midrule
\multicolumn{4}{l}{\textbf{Zero Shot Prompting}} \\
\midrule
ChatGPT3.5 & 0.72 & 0.53 & 0.72 & 0.49 \\
ChatGPT4 & 0.59 & 0.52 & 0.59 & 0.50 \\
CodeLlama-7b-Instruct-hf & 0.42 & 0.33 & 0.28 & 0.13 \\
StackLlama & 0.40 & 0.36 & 0.26 & 0.06 \\
\midrule
\multicolumn{4}{l}{\textbf{Prompt Engineering}} \\
\midrule

ChatGPT3.5@10+CoT & 0.78 & 0.55 & 0.78 & 0.57 \\
ChatGPT4@10+CoT & {0.59} & 0.51 & {0.59} & 0.50 \\

\midrule
\multicolumn{4}{l}{\textbf{Fine-tuning}} \\
\midrule

StackLlama & 0.72 & \textbf{0.85} & 0.72 & 0.77 \\
CodeLlama-7b-Instruct-hf & 0.76 & 0.83 & 0.76 & 0.79 \\

\midrule
\multicolumn{4}{l}{\textbf{Pre-trained BERT}} \\
\midrule
BertOverflow & 0.49 & 0.47 & 0.49 & 0.48 \\
BBU & 0.70 & 0.70 & 0.70 & 0.70 \\
BLU & 0.48 & 0.47 & 0.48 & 0.47 \\
SOBertBase & \textbf{0.92} & {0.76} & \textbf{0.92} & \textbf{0.82} \\
SOBertLarge & {0.71} & {0.81} & {0.71} & {0.75} \\
\bottomrule
\end{tabularx}
\caption{Performance Metrics for the Models for Obsoletion Detection {\includegraphics[height=1em]{sections/Images/4.png}}}
\vspace{-20pt}
\label{tab:ComparisonChatGpt}
\end{table}
While this runs counter to the wisdom that larger language models are better \cite{kaplan2020scaling}, we note that this notion typically applies to models that are not fine-tuned but rather prompted (queried). Fine-tuning adds the noteworthy constraint that larger models represent their inputs in more hidden dimensions, which means that the task-specific prediction layer(s) also require more parameters. For instance, predicting a 1-dimensional label from a 768-dimensional hidden state (for the Base models) requires training half as many new parameters as transforming the 1,536-dimensional states from the Large models. Tuning more parameters also increases the risk of overfitting as models can quickly memorize small datasets.


\added{A noteworthy observation from the training process was that, while loss continued to decline (\Cref{fig:lossPlotsComparison}), downstream task performance saturated around halfway through training, at approximately 50,000 steps, for both SOBertBase and SOBertLarge. While language models generally continue to improve on end tasks as the training loss decreases, this trend does not necessarily hold for models that involve a separate task-specific fine-tuning phase. The additional fine-tuning step appears to enable BERT architectures to reach peak downstream performance well before full convergence of the base pre-trained model.

This highlights that in-domain BERT models need not be exhaustively pre-trained in order to realize most of the gains from domain-specific optimization. Had we halted training at 50,000 steps, we would have incurred only about half of the total compute cost, while likely retaining the bulk of the performance gains over general-domain BERTs demonstrated in this work. This insight could impart significant cost savings when pre-training models for new domains.}

\subsection{Relation to LLMs} 

The results indicate that, while LLMs can solve many NLP problems quite well, they do not fare well on the tasks studied in this work. A potential contributing factor is that our downstream tasks are all classification tasks, whereas nearly all LLMs are decoder-only, causal language models, which lend themselves especially to generative tasks. BERT-like models, on the other hand, are built upon a bidirectional Masked Language Modeling (MLM) objective that emphasizes the encoding of contextual information. Another, perhaps more significant factor is the domain-specific and challenging nature of the tasks studied here. Predicting question quality, for instance, is quite challenging even for a human reader (for some examples, see the few-shot prompt in the appendix). Mastering this task evidently requires exposure to a large number of labeled examples. Even 10-shot prompting GPT-4 does not meaningfully shrink the performance gap with SOBert, which sees 60K training examples in each epoch. While LLMs can be fine-tuned, it is impractical to do so for every downstream task. As the results with Llama models in \Cref{tab:ComparisonChatGpt} show, this also does not necessarily bridge the gap with purpose-tuned bidirectional models.

One might think that waiting for a newer and bigger LLM to be released could improve performance. However, we see that GPT-4 shows lower performance compared to GPT-3.5 in our experiments, suggesting that simply relying on the next generation of LLMs may not be the solution for such domain-specific tasks. These findings underscore the importance of carefully considering the task at hand and the available data before choosing the appropriate model, rather than defaulting to the latest LLM.

\section{Conclusion}

In this study, we show that small, domain-specific language models built by combining SE insights with ML best-practices can yield superior results to generalist LLMs and BERT models. We trained SOBertBase and SOBertLarge on SO data with hyper-parameter configurations tailored towards the data characteristics of SO posts. With 125M and 762M parameters respectively, these models were trained with a modest budget of \$374 and \$1600. Our models consistently outperformed baseline models across four software engineering-related tasks: question quality prediction, closed question prediction, named entity recognition, and obsoletion detection. We introduce obsoletion detection as a new benchmark task. We compare our models with a range of other BERT models and with LLMs such as GPT-3.5, GPT-4, CodeLlama, and StackLlama. Despite their comparatively smaller size, the SOBert models consistently outperformed all other models, emphasizing the importance of task-specific model training.
Our study demonstrates that for certain tasks, developing tailored, domain-specific models can be a powerful and cost-effective alternative to relying solely on large, general-purpose language models.

\section{DATA AVAILABILITY}
Our supplemental materials are available on Figshare \cite{figshare}.
It includes the newly introduced obsolete answers dataset,
our annotation protocol, all prompts used and the weights for both SOBert models.


\bibliographystyle{IEEEtranN}
\bibliography{bibliography}


\end{document}